\documentclass[sigconf]{acmart}

\AtBeginDocument{%
  }

\setcopyright{acmlicensed}
\copyrightyear{2018}
\acmYear{2018}
\acmDOI{XXXXXXX.XXXXXXX}

\acmConference[Conference acronym 'XX]{Make sure to enter the correct
  conference title from your rights confirmation emai}{June 03--05,
  2018}{Woodstock, NY}
\acmISBN{978-1-4503-XXXX-X/18/06}

\usepackage{xspace}
\usepackage{enumitem}
\usepackage{amsmath}
\usepackage{pifont}
\newcommand{\method}{\textsc{Hound}\xspace}
\newcommand{\stitle}[1]{\vspace*{0.4em}\noindent{\bf #1\/}}
\usepackage{multirow}
\usepackage[normalem]{ulem}
\useunder{\uline}{\ul}{}
\usepackage{setspace}
\usepackage{url}
\newcommand{\squishlist}{
	\begin{list}{$\bullet$}
		{ \setlength{\itemsep}{1pt}
			\setlength{\parsep}{1pt}
			\setlength{\topsep}{2.5pt}
			\setlength{\partopsep}{0.5pt}
			\setlength{\leftmargin}{1em}
			\setlength{\labelwidth}{1em}
			\setlength{\labelsep}{0.6em}
		}
	}
	\newcommand{\squishend}{
	\end{list}
}

\begin{document}

\title{Hound: Hunting Supervision Signals for Few and Zero Shot \\ Node Classification on Text-attributed Graph}

\author{Yuxiang Wang}
\affiliation{%
  \institution{School of Computer Science,\\ Wuhan University}
  \city{Wuhan}
  \country{China}
}
\email{nai.yxwang@whu.edu.cn}

\author{Xiao Yan}
\affiliation{%
  \institution{Centre for Perceptual and Interactive Intelligence (CPII)}
  \city{Hong Kong}
  \country{China}}
\email{yanxiaosunny@gmail.com}

\author{Shiyu Jin}
\affiliation{%
  \institution{School of Computer Science,\\ Wuhan University}
  \city{Wuhan}
  \country{China}
}
\email{syjin@whu.edu.cn}

\author{Quanqing Xu}
\affiliation{%
  \institution{OceanBase, Ant Group}
  \city{Hangzhou}
  \country{China}
}
\email{xuquanqing.xqq@oceanbase.com}

\author{Chuanhui Yang}
\affiliation{%
  \institution{OceanBase, Ant Group}
  \city{Hangzhou}
  \country{China}
}
\email{rizhao.ych@oceanbase.com}

\author{Chuang Hu}
\affiliation{%
  \institution{School of Computer Science,\\ Wuhan University}
  \city{Wuhan}
  \country{China}
}
\email{handc@whu.edu.cn}

\author{Yuanyuan Zhu}
\affiliation{%
  \institution{School of Computer Science,\\ Wuhan University}
  \city{Wuhan}
  \country{China}
}
\email{yyzhu@whu.edu.cn}

\author{Bo Du}
\affiliation{%
  \institution{School of Computer Science,\\ Wuhan University}
  \city{Wuhan}
  \country{China}
}
\email{dubo@whu.edu.cn}

\author{Jiawei Jiang}
\affiliation{%
  \institution{School of Computer Science,\\ Wuhan University}
  \city{Wuhan}
  \country{China}
}
\email{jiawei.jiang@whu.edu.cn}

\renewcommand{\shortauthors}{Trovato et al.}

\begin{abstract}
Text-attributed graph (TAG) is an important type of graph structured data with text descriptions for each node. Few- and zero-shot node classification on TAGs have many applications in fields such as academia and social networks. However, the two tasks are challenging due to the lack of supervision signals, and existing methods only use the contrastive loss to align graph-based node embedding and language-based text embedding. In this paper, we propose \method to improve accuracy by introducing more supervision signals, and the core idea is to go beyond the node-text pairs that come with data. Specifically, we design three augmentation techniques, i.e., \textit{node perturbation}, \textit{text matching}, and \textit{semantics negation} to provide more reference nodes for each text and vice versa. Node perturbation adds/drops edges to produce diversified node embeddings that can be matched with a text. Text matching retrieves texts with similar embeddings to match with a node. Semantics negation uses a negative prompt to construct a negative text with the opposite semantics, which is contrasted with the original node and text. We evaluate \method on 5 datasets and compare with 13 state-of-the-art baselines. The results show that \method consistently outperforms all baselines, and its accuracy improvements over the best-performing baseline are usually over 5\%. 

\end{abstract}





\maketitle

\begin{figure*}[t]
    \centering
    \includegraphics[scale=0.5]{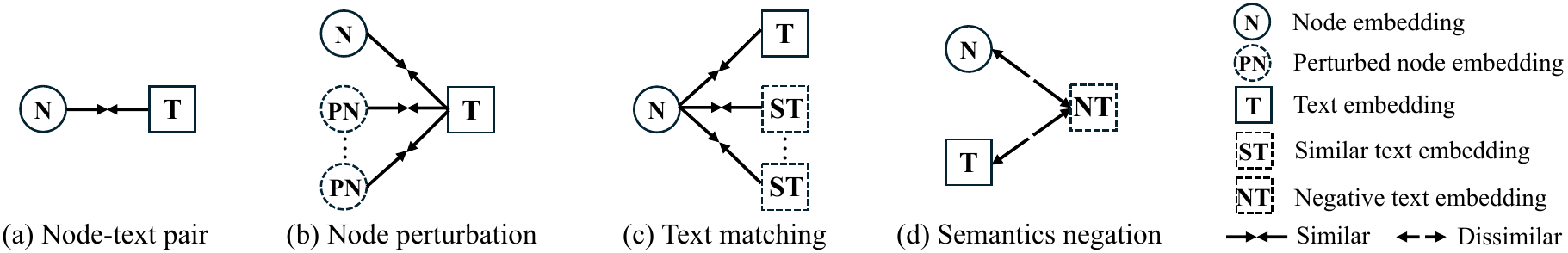}
    \caption{The contrastive loss of G2P2 (a) and three supervision signals proposed by \method (b-d). The node-text pair of G2P2 is specified by data as each node has a text description, and \method provides more reference nodes for each text and vice versa.}
    \label{fig:alpha}
\end{figure*}

\section{Introduction}
Text-attributed graph (TAG) \cite{yang2021graphformers,zhang2020text,zhao2022learning} is a prevalent type of graph-structured data, where each node is associated with a text description. For instance, in a citation network, the papers (i.e., nodes) are linked by the citation relations (i.e., edges), and the abstract of each paper serves as the text description. Few-shot and zero-shot node classification on TAGs (FZNC-TAG) predict the categories of the nodes using a few or even no labeled data since labeled data are expensive to obtain ~\cite{liu2021relative,liu2022few,wang2021zero,wang2022task,zhou2019meta}. The two tasks have many applications in areas such as recommender system~\cite{gao2022graph,wang2019knowledge}, social network analysis~\cite{yu2020enhancing,zhao2021community}, and anomaly detection~\cite{noble2003graph,deng2021graph}. 

Existing methods for FZNC-TAG typically follow a two-step process: first learn node and text embeddings on the TAGs and then use prompting to produce classification results~\cite{wen2023augmenting, huang2023prompt}. They mainly differ in embedding learning and can be classified into three categories. \ding{182} Some methods use pre-trained language models (PLMs) such as Bert~\cite{devlin2018bert}, RoBERTa~\cite{liu2019roberta}, and GPT~\cite{radford2018improving} to generate text embeddings. They exploit the text but ignores the information in the graph topology. \ding{183} Some methods encode the text using PLMs and add the text embedding as additional node features. Then, semi-supervised graph neural network (GNN) methods, e.g., TextGCN~\cite{yao2019graph} and GraphSAGE~\cite{hamilton2017inductive}, are used to train the node embeddings. The limitation of these approaches is that the PLMs are not updated during GNN training~\cite{yan2023comprehensive}. \ding{184} The state-of-the-art method, G2P2~\cite{wen2023augmenting}, jointly trains the GNN and language model via self-supervised learning. As shown in Figure~\ref{fig:alpha}(a), G2P2 uses a popular contrastive loss~\cite{he2020momentum} to ensure that the GNN and PLMs produce similar embeddings for each node-text pair.

However, we observe that the classification accuracy of G2P2 is still low. For instance, on the Fitness dataset~\cite{yan2023comprehensive}, G2P2 only achieves an accuracy of 68.24\% and 45.99\% for few- and zero-shot classification, respectively. The significantly lower accuracy of zero-shot classification suggests that the problems are caused by the lack of supervision signals, and that the contrastive loss employed by G2P2 is inadequate for training high-quality models. Thus, we ask the following research question:
\begingroup
\addtolength\leftmargin{0.001in}
\addtolength\rightmargin{0.001in}
\begin{quote}
    \textit{How to provide more supervision signals to enhance training for few- and zero-shot classification on TAGs?}
\end{quote}
\endgroup

To answer the above question, we propose \method, which enhances supervision signals by mining more node-text pairs. In particular, embedding learning can be improved by enforcing similarity relations between the embeddings. As shown in Figure~\ref{fig:alpha}(a), G2P2 is limited to make the node-text pairs in the TAGs similar. We can create additional node-text pairs that should have similar or dissimilar embeddings to facilitate model learning. Using the idea, we design three augmentation techniques, which are illustrated in Figure~\ref{fig:alpha}(b-d) and elaborated as follows.

\stitle{Node perturbation.} In Figure~\ref{fig:alpha}(b), we provide multiple node embeddings for each text embedding. This is achieved by randomly adding or removing a portion of edges in the graph such that the GNN produces different embeddings for the same node. We encourage the text embedding to be similar to these perturbed node embeddings. This argumentation makes the text encoder (i.e., language model) robust to graph topology and enforces that minor changes in topology should not change classification results.

\stitle{Text matching.} In Figure \ref{fig:alpha}(c), we provide multiple text embeddings for each node embedding. This is achieved by searching the texts that have similar embeddings to the text of the considered graph node. We also encourage the node embedding to be similar to those text embeddings. This augmentation provides more supervision to GNN training and enforces the prior that nodes may belong to the same categories if their texts have similar meanings.

\stitle{Semantic negation.} In Figure \ref{fig:alpha}(d), we pair each text with a semantically opposite \textit{negative text} by adding some learnable tokens to the front of the text. We then encourage the embeddings of the original text and its corresponding node to be dissimilar to the negative text. This argumentation provides additional semantics supervisions to make the classification robust. For instance, to classify a paper as being related to data mining, it should not only be similar to the description \textit{``a paper is published at KDD''} but also be dissimilar to \textit{``a paper is not published at KDD''}.

We conduct extensive experiments to evaluate \method, using 5 datasets and comparing with 13 state-of-the-art baselines. The results show that \method consistently achieves higher accuracy than all baselines across the datasets and for both  few-shot and zero-shot classification. In particular, \method improves the accuracy and F1 scores of few-shot classification by 4.6\% and 6.9\% on average, and zero-shot classification by 8.8\% and 9.3\%, respectively. Ablation study shows that all our augmentation techniques improve accuracy. Moreover, timing experiments show that the augmentation techniques do not incur significant overheads, and the running time of \method is comparable to the state-of-the-art baselines.

In summary, we make the following contributions:
\begin{itemize}[leftmargin=*]
    \item We observe that existing methods for few- and zero-shot node classification on text-attributed graph suffer from the lack of supervision signals and thus have poor accuracy.
    \item  We propose \method to provide more supervision signals by generating more node-text pairs for training, and the idea may be extend beyond our augmentation techniques.
    \item We design three augmentation techniques, i.e., node perturbation, text matching, and semantics negation, for mining supervision signals from both the graph and text modalities.
    \item We conduct extensive experiments to evaluate \method, validating its good accuracy and efficiency.
\end{itemize}

\section{Preliminaries}
\stitle{Text-attributed graph.} We denote a text-attributed graph (TAG) as $\mathbf{G}=(\mathcal{V},\mathcal{E},\mathbf{X})$, in which $\mathcal{V}$, $\mathcal{E}$, and $\mathbf{X}$ are the node set, edge set, and text set, respectively. Take citation network as an example for TAG. Each node $v_i\in \mathcal{V}$ is a paper, interconnected by the edges $e\in \mathcal{E}$ that signify citation relations. Let $\mathbf{x}_{i}\in \mathbf{X}$ denote the text description (i.e., paper abstract) of the $i$-th node. Each node has a ground-truth label to indicate the topic of the paper. Since the graph nodes and papers have a strict one-to-one correspondence, node $v_{i}$ and text $\mathbf{x}_i$ share an identical label.

\stitle{Few- and zero-shot learning.} For few-shot classification, the test dataset encompasses a support set $\mathcal{S}$ and a query set $\mathcal{Q}$. $\mathcal{S}$ comprises $C$ classes of nodes, with $K$ labeled nodes drawn from each class. These nodes can be used to train or fine-tune the classifier, which is then utilized to classify the nodes in $\mathcal{Q}$. This is also called the $C$-way $K$-shot classification problem~\cite{finn2017model}. Zero-shot node classification is essentially a special case of few-shot classification with $K=0$. There are no labeled nodes for both  training and testing, and classification depends solely on the class names.

\stitle{Contrastive loss.} Recent researches \cite{wen2023augmenting,zhao2024pre} use the contrastive loss to jointly train the graph and text encoders. Specifically, they employ GNNs~\cite{kipf2016semi} as the graph encoder $\phi$ to encode each node $v_i$ into a node embedding $\mathbf{n}_i$, and adopt Transformer~\cite{vaswani2017attention} as the text encoder $\psi$ to map each text $\mathbf{x}_i$ to  a text embedding $\mathbf{t}_i$. That is,
\begin{equation}
    \mathbf{n}_i=\phi (v_i), \quad \mathbf{t}_{i}=\psi (\mathbf{x}_i).
\end{equation}

\begin{figure*}[t]
    \centering
    \includegraphics[scale=0.6]{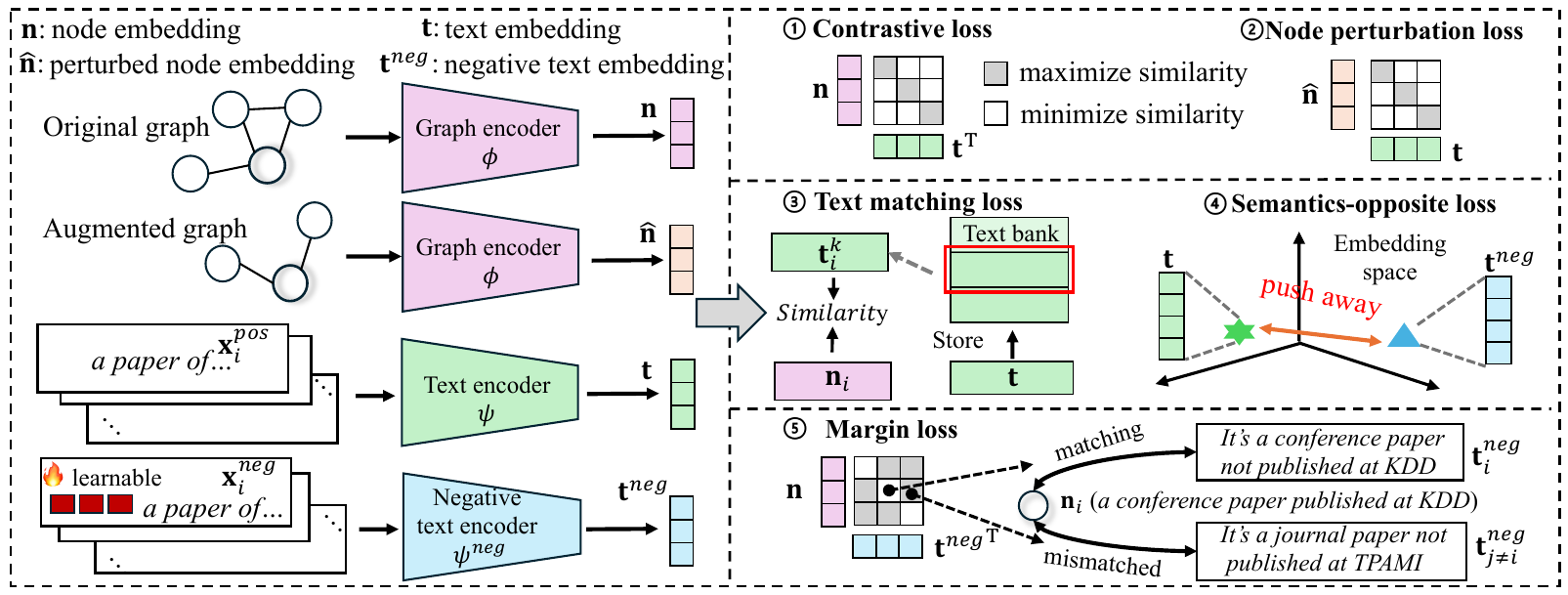}
    \caption{The overview of \method.}
    \label{fig:framework overview}
    \vspace{-0.2cm}
\end{figure*}

Then, they use InfoNCE~\cite{he2020momentum} loss $\mathcal{L}_{CL}$ to maximize the similarity between each node $\mathbf{n}_{i}$ and its corresponding text $\mathbf{t}_{i}$, while simultaneously minimizing the similarity between node $\mathbf{n}_{i}$ and other mismatched texts $\mathbf{t}_{j}$.  As shown in part (1)  of Figure~\ref{fig:framework overview}, $\mathcal{L}_{CL}$ is calculated as follows:
\begin{equation}
    \mathcal{L}_{CL}= -\frac{1}{\left | \mathcal{B} \right | } \!\sum_{(\mathbf{n}_{i},\mathbf{t}_{i})\in \mathcal{B}}^{} \!\mathrm{log} \frac{\mathrm{exp}(\mathrm{sim}(\mathbf{n}_{i}, \mathbf{t}_{i})/\tau) } { {\textstyle \sum_{j\ne i }}  \mathrm{exp}(\mathrm{sim}(\mathbf{n}_{i},\mathbf{t}_{j})/\tau) },
\label{eq:cl loss}
\end{equation}
where $\mathcal{B}$ is a batch of training instances, $\mathrm{sim}(,)$ is the cosine similarity, and $\tau$ is a learnable temperature. In essence, the objective aims to align the embeddings of each node-text pair in the data.


\section{The \method Framework}
In this section, we present a novel pre-training and prompting framework, named \method, designed for the TAGs under the few- and zero-shot setting. We start with a framework overview and follow up with the detailed descriptions of its components.

\subsection{Overview}
The overall architecture of our framework is illustrated in Figure~\ref{fig:framework overview}. The pre-training model for few-shot consists of a graph encoder and a text encoder, and an extra negative text encoder is included for zero-shot pre-training. We introduce them as follows.
\begin{itemize}[leftmargin=*]
    \item \textbf{Graph encoder} $\phi$. We adopt a graph neural network as the encoder to generate the node embedding $\mathbf{n}$.
    \item \textbf{Text encoder} $\psi$. We choose Transformer~\cite{vaswani2017attention} as the text encoder, and it produces a text embedding $\mathbf{t}$ for each text description.
    \item \textbf{Negative text encoder} $\psi^{neg}$. This maintains the same architecture and inputs as the text encoder, with the difference that we train it independently with a negative prompt to generate the negative text presentation $\mathbf{t}^{neg}$.
\end{itemize}

To effectively train the above encoders, we design three novel loss functions: \textit{node perturbation loss}, \textit{text matching loss}, and \textit{semantics negation loss}, which can assist the pre-training model in acquiring more supervision signals.
Then, we detail the basic paradigm for few- and zero-shot node classification. Finally, we propose a strategy in Section~\ref{sec:inference strategy}, \textit{probability-average}, to enhance zero-shot node classification by merging the probabilities produced from both the text encoder and the negative text encoder outputs.

\subsection{Supervision Signals}
\label{sec: ssl pretrain model}
The current methods~\cite{wen2023augmenting,zhao2024pre,huang2023prompt} neglect supervision signals within the graph and text modalities during the pre-training phase. Therefore, in this section, we introduce three novel augmentation techniques: node perturbation, text matching, and semantics negation, to provide more nodes for each text and vice versa.

\stitle{Node perturbation loss.} The prior research~\cite{wen2023augmenting} solely contrasts the original node (without perturbation) with text $\mathbf{t}$ (i.e., Equation \eqref{eq:cl loss}). However, it fails to fully exploit the supervision signals in the graph modality. To solve this issue, as shown in Figure~\ref{fig:framework overview}(2), we propose node perturbation to introduce more nodes for text embeddings. Specifically, we generate multiple perturbed nodes by randomly removing or adding a portion of edges, and then maximize the similarity between the text embedding $\mathbf{t}_{i}$ and the perturbed node embedding $\hat{\mathbf{n}}_{i}$. The rationale behind the augmentation technique is that when the node perturbation is applied, the corresponding prior of the node data distribution is injected, forcing the text encoder to learn an embedding that is invariant to the node perturbation. The benefits of this are intuitive: first, the model does not change the classification results due to minor changes in topology; second, the node perturbations provide the text with more pairs of samples, facilitating the text to learn a more generalized embedding. The node perturbation loss can be represented as follows:
\begin{equation}
    \mathcal{L}_{NP}=-\frac{1}{\left | \mathcal{B} \right | }\!\sum_{(\hat{\mathbf{n}}_{i},\mathbf{t}_{i})\in \mathcal{B}}\!\mathrm{log} \frac{\mathrm{exp}(\mathrm{sim}(\hat{\mathbf{n}}_{i},\mathbf{t}_{i})/\tau) } { {\textstyle \sum_{j\ne i}}  \mathrm{exp}(\mathrm{sim}(\hat{\mathbf{n}}_{i},\mathbf{t}_{j})/\tau) },
    \label{eq:node perturbation loss}
\end{equation}
where $\hat{\mathbf{n}}_{i} = \phi (\zeta (v_{i}))$ is generated by the graph encoder with a perturbed node as input, and $\zeta()$ is the augmentation function.

Note that a similar data augmentation operation is performed in graph contrastive learning~\cite{you2020graph,zhu2021graph}. It is used to generate diverse nodes and mitigate over-smoothing resulting from the shallower GNN structure. Differently, our objective is to provide more perturbed nodes for the texts to address the lack of supervision signals in the few- and zero-shot classification. Thus, the application scenarios and purposes of these two approaches differ significantly.


\stitle{Text matching loss.} In addition to providing multiple node embeddings for text embeddings, in turn, we also provide more text embeddings for each node embedding. G2P2~\cite{wen2023augmenting} defaults to only one text embedding similar to each node embedding, however there may be multiple similar texts to the target node in TAGs~\cite{he2020momentum,chuang2020debiased,zhang2022m}. Therefore, we search for multiple text embeddings that are similar to the text embedding of the target node and subsequently encourage the target node embedding to align with these similar text embeddings $\hat{\mathbf{t}}$. The text matching loss is denoted as follow:
\begin{equation}
    \mathcal{L}_{TM}=-\frac{1}{\left | \mathcal{B} \right | }\!\sum_{(\mathbf{n}_{i},\mathbf{t}_{i})\in \mathcal{B}}\!\mathrm{log} \frac{ {\textstyle \sum_{k=1}^{\mathbf{K}}} \mathrm{exp}(\mathrm{sim}(\mathbf{n}_i ,\hat{\mathbf{t}}_{i}^{k} )/\tau) } { {\textstyle \sum_{j\ne i}}  \mathrm{exp}(\mathrm{sim}(\mathbf{n}_{i},\mathbf{t}_{j})/\tau) },
    \label{eq:text matching loss}
\end{equation}
where $\mathbf{K}$ is the number of similar text embeddings.

However, the above method has two serious drawbacks: first, the complexity of violently searching for similar text embeddings among all text embeddings is unacceptable; second, storing all text embeddings in GPU memory may lead to out-of-memory. To address these issues, we create a text bank with a capacity of 32K to model the whole text embedding space. As illustrated in the Figure~\ref{fig:framework overview}(3), whenever a new batch of data arrives, the earliest text embedding is discarded if the capacity of the text bank exceeds a predetermined threshold. Otherwise, it is stored in the bank. Subsequently, we identify the most $\mathbf{K}$ similar text embeddings for target node through similarity calculations. In this way, our text bank is both time-efficient and space-efficient.

\stitle{Semantics negation.} After co-training the graph and text encoder using Equations~\eqref{eq:cl loss}, \eqref{eq:node perturbation loss} and \eqref{eq:text matching loss}, the model now possesses the base capability to distinguish node-text pairs. However, understanding the negative semantics within the input text description poses a challenge for the model. For example, we represent a text description such as \textit{``a paper is published at KDD''} and its negation \textit{``a paper is not published at KDD''}. In the embedding space, these two descriptions are likely to be very similar, as their raw texts differ by only one word. To address this issue, we employ negative prompts to generate multiple negative texts that are semantically opposed to the original text descriptions. These negative texts are then used to train a negative text encoder independently. This process helps the negative text encoder learn parameters that are contrary to those of the text encoder. This augmentation technique generates an additional negative text for the nodes and texts, providing more semantics supervisions to make the classification robust.
Next we detail the design of negative prompts and negative text encoders.

Our initial idea is to manually construct a series of negative texts. Specifically, we alter the text descriptions by incorporating negation terms such as \textit{``no'', ``not'', ``without''}, etc., thus creating a negative text corpus that are semantically opposite to the original ones, denoted as $\mathbf{X}^{neg}$. Then, we input the negative text $\mathbf{x}_{i}^{neg}$ into negative text encoder $\psi^{neg}$ to generate negative text embedding $\mathbf{t}^{neg}_i$, as denoted below:
\begin{equation}
    \mathbf{t}^{neg}_i = \psi^{neg}(\mathbf{x}_{i}^{neg}), \quad \mathbf{x}^{neg}\in \mathbf{X}^{neg}.
\end{equation}

However, manual modification of the raw text is time-consuming and labor-intensive. To solve this problem, inspired by CoOp~\cite{zhou2022conditional}, we propose a \textit{learnable negative prompt} $\mathbf{h}$ and add it to the front of raw text. The underlying logic is to represent negative semantics by constantly optimizing the learnable $\mathbf{h}$, thereby mirroring the hand-crafted negative texts. Specifically, we splice the text description with $\mathrm{M}$ learnable vectors and then input it into the negative text encoder $\psi^{neg}$, denoted as follows:
\begin{equation}
    \mathbf{h}=[\underbrace{\mathrm{V}_1, \mathrm{V}_2,... \mathrm{V}_{\mathrm{M} },}_{\mathrm{negative\ prompt}}\mathbf{x} ], \quad \mathbf{t}^{neg}_i = \psi^{neg}(\mathbf{h}^{neg}_i),
    \label{eq:learnable negative prompt}
\end{equation}
where the negative text encoder is a transformer~\cite{vaswani2017attention} with the same architecture as the text encoder.

There is still an unsolved problem: \textit{how do we train a negative text encoder?} In other words, how do we ensure that the semantics of the negative text embeddings contradict the original text embeddings. To address this, we design two novel loss functions: \textit{margin loss} and \textit{semantics-opposite loss}.

The margin loss anticipates the greatest possible similarity between positive pairs, and conversely, it expects dissimilarity in the case of negative pairs.
As shown in Figure~\ref{fig:framework overview}(5), given a target node $v_i$ (i.e., \textit{``a conference paper published at KDD''}), the corresponding negative text description $\mathbf{t}^{neg}_i$ (i.e., \textit{``it's not a conference paper published at KDD''}) is deemed a negative text, while any other non-corresponding text $\mathbf{t}^{neg}_{j\ne i}$ (i.e., \textit{``it's not a journal paper published at TPAMI''}) are considered positive texts. Subsequently, we employ margin loss to assess the degree of matching between the target nodes, positive texts, and negative texts. Specifically, margin loss ensures that the similarity between the target node embedding and the positive text embedding is at least a margin higher than the similarity with the negative text embedding. We maintain a margin of up to $m$ with no additional benefits for further widening this gap. The margin loss $\mathcal{L}_{ML}$ is denoted as follows:
\begin{equation}
    \mathcal{L}_{ML}= max(0, m+\text{sim}(\mathbf{n}_i,\mathbf{t}^{neg}_{i})-\text{sim}(\mathbf{n}_i,\mathbf{t}^{neg}_{j\ne i} ))
    \label{eq:ml loss}
\end{equation}

As shown in Figure~\ref{fig:framework overview}(4), semantics-opposite loss seeks to maximize the mean square error between positive and negative text embeddings. As text $\mathbf{x}_i$ and negative text $\mathbf{t}^{neg}_i$ are semantically opposite, their corresponding embeddings should be as far apart as possible in the text embedding space. We compute the semantics-opposite loss as follow:
\begin{equation}
    \mathcal{L}_{SO}=-\frac{1}{\left | \mathcal{B} \right | }\sum_{\mathbf{t}_{i}\in \mathcal{B}}\left \| \mathbf{t}_{i}- \mathbf{t}_{i}^{neg} \right \|_{2},
\end{equation}
where $\left \|  \right \| _2$ is the L2 norm. Thus, the semantics negation loss is equal to the sum of margin loss and semantics-opposite loss, denoted as $\mathcal{L}_{SN}=\mathcal{L}_{ML}+\mathcal{L}_{SO}$. It enforces both the node and text embeddings are dissimilar to the corresponding negative text embedding.

In summary, we denote the total loss of our \method as:
\begin{equation}
    \mathcal{L}= \mathcal{L}_{CL}+\alpha \mathcal{L}_{NP}+\beta \mathcal{L}_{TM}+  \gamma \mathcal{L}_{SN},
\end{equation}
where $\alpha$, $\beta$ and $\gamma$ are the hyperparameters used to balance the loss. 
In few-shot pre-training, we do not activate the semantic negation loss (i.e., $\gamma=0$) because the prompts in few-shot are inherently learnable, incorporating negative prompts would introduce more noise and lead to sub-optimal performance. In contrast, zero-shot classification lacks labeled data during the pre-training. Thus, we activate the semantics negation loss to provide more supervision signals (i.e., $\gamma=1$). Overall, in the total loss $\mathcal{L}$, we only modify $\alpha$ and $\beta$, which does not require extensive hyperparameter tuning.
We analyze the ablation experiments on the loss function in detail in Section~\ref{sec:ablation study}.

\stitle{Complexity Analysis.} Our pre-trained model incorporates both a GNN and a Transformer. The GNN takes $O(LNd^2)$ time for aggregating the neighboring nodes, where $L$ is the network depth, $N$ is the number of nodes and $d$ is the number of dimensions. The Transformer's time complexity is $O(sd^2+s^2d)$, where $s$ the maximum length of the input sequence. $O(sd^2)$ time is used for mapping vectors at each position to query, key and value vectors, and $O(s^2d)$ time is utilized for the computation of the attention score. Consequently, the overall time complexity of our method is $O(LNd^2+sd^2+s^2d)$. The state-of-the-art G2P2 also contains a GNN and Transformer, so the time complexity of our method is comparable to its. In this paper, computations during pre-training are performed in batches, where the number of batch $|\mathcal{B}|<<N$. Thus the actual complexity is significantly lower than $O(LNd^2+sd^2+s^2d)$. 

\subsection{Prompt Tuning and Inference}
\label{sec:inference strategy}
Based on the pre-trained model, we tune the model parameters to adapt to the classification tasks. However, there are two limitations inherent in the conventional pre-training and fine-tuning paradigm~\cite{sun2022gppt,sun2023all,liu2023graphprompt}. Firstly, it requires labeled data for training a prediction head, such as an MLP layer. Secondly, it requires fine-tuning the enormous parameters of pre-trained model. These issues can be solved through few- and zero-shot learning, which enables classification with a few even no labeled samples while concurrently freezing the pre-trained model's parameters. Next, we introduce the foundational paradigm of few- and zero-shot node classification. 

\stitle{Zero-shot classification.} In the zero-shot setting, we operate without any labeled samples and rely solely on class name description. To perform $C$-way node classification, we construct a series of class descriptions $\{\mathbf{D}_c\}_{c=1}^{C}$, via discrete prompts, such as ``\textit{a paper of} [\textit{class}]''. Then, we input the description text into the pre-trained text encoder to generate the class embedding $\mathbf{g}_c=\psi (\mathbf{D}_c)$.
We predict the category of a node $v_i$ by computing the similarity between the node embedding $\mathbf{n}_i$ with the class embedding $\mathbf{g}_c$. The insight behind this is that we align the pre-training and prompting objectives (i.e., to determine whether nodes and texts are similar). Thus, we do not have to tune the parameters of the pre-trained model. The similarity probability between the target node and the candidate class description is calculated as follows:
\begin{equation}
    p_{i}=\frac{\text{exp}(\text{sim}(\mathbf{n}_i,\mathbf{g}_c) /\tau) }{ {\textstyle \sum_{c=1}^{C}\text{exp}(\text{sim}(\mathbf{n}_i,\mathbf{g}_c)/\tau ) } }.
    \label{eq:positive classification probability}
\end{equation}

\begin{figure}[t]
    \centering
    \includegraphics[scale=0.5]{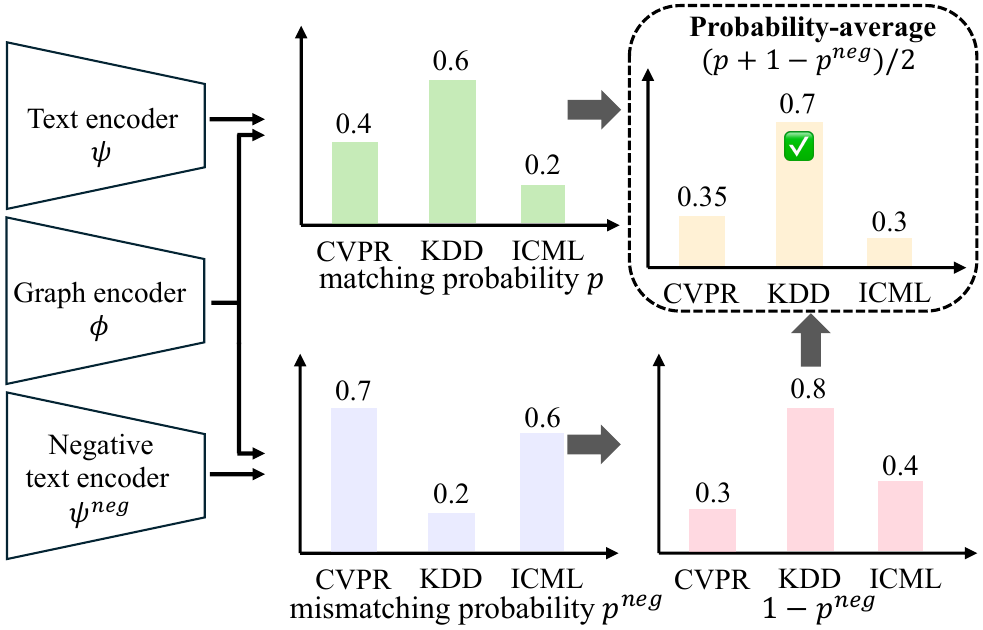}
    \vspace{-0.2cm}
    \caption{The illustration of probability-average.}
    \label{fig:inference stratery}
    \vspace{-0.4cm}
\end{figure}

\stitle{Few-shot classification.} In the few-shot setting, we conduct a $C$-way $K$-shot classification task. Unlike discrete prompts (i.e., ``\textit{a paper of ...}'') in the zero-shot setting, we have $C\times K$ labeled samples to train learnable prompts.
Specifically, we construct a continuous prompt $\mathbf{g}_c$ by adding $M$ learnable vectors to the front of the class description $\mathbf{D}_c$. Formally, we denote $\mathbf{g}_c=\psi ([\mathbf{e}_1, \mathbf{e}_2,...,\mathbf{e}_M,\mathbf{D}_c ] )$. Then, we use Equation~\eqref{eq:positive classification probability} to predict the node category, and update the continuous prompts by minimizing the discrepancy between the predicted and ground-truth labels via cross-entropy loss. It is worth noting that because $C\times K$ is a small value, the parameters required to fine-tune the prompts are considerably less than those needed for the pre-trained model. 

\stitle{Probability-average.} As shown in Figure~\ref{fig:inference stratery}, we propose probability-average to predict node category. Specifically, we first compute $p_{i}$ by Equation ~\eqref{eq:positive classification probability}. We use the negative text encoder to generate the negative class embedding. Then, we compute negative probability $p^{neg}_{i}$ by contrasting these negative class embeddings with the target node embedding using Equation~\eqref{eq:positive classification probability}. $p_{i}$ denotes the probability that a node belongs to each category and vice versa, $p^{neg}_{i}$ represents the probability that a node does not belong to each category. Finally, we utilize $(p_{i}+1-p^{neg}_{i})/2$ to predict the node label. Unlike using a single text encoder, integrating a negative text encoder provides additional evaluation metric. This strategy balances the output probabilities of the positive and negative text encoders, thereby enhancing classification accuracy. Formally, the probability-average strategy can be denoted as follows:
\begin{equation}
    \mathcal{Y}_i =\text{arg} \ \text{max}\ (p_i +1-p_i^{neg})/2.
    \label{eq:probability-average}
\end{equation}

Note that the probability-average stragety is only applicable to zero-shot classification, as it requires the negative prompts and negative text encoder to calculate $p^{neg}_{i}$. In contrast, few-shot classification directly uses $p_{i}$ to predict node categories. This is because few-shot classification can learn a prompt from labeled samples, and the additional introduction of negative prompts introduces noise and may reduce accuracy.

\begin{table}[t]
\caption{Statistics of the experiment datasets.}
\vspace{-0.2cm}
\label{tab:dataset statistic}
\begin{tabular}{c|cccc}
\bottomrule
\textbf{Dataset} & \textbf{\# Nodes} & \textbf{\# Edges} & \textbf{\# Avg.deg} & \textbf{\# Classes} \\
\midrule
Cora             & 25,120           & 182,280          & 7.26                 & 70                 \\
Fitness          & 173,055          & 1,773,500        & 17.45                 & 13   \\
M.I.             & 905,453          & 2,692,734        & 2.97                 & 1,191              \\
Industrial       & 1,260,053        & 3,101,670        & 2.46                 & 2,462              \\
Art              & 1,615,902        & 4,898,218        & 3.03                 & 3,347              \\
\bottomrule
\end{tabular}
\end{table}

\section{Experimental Evaluation}
In this section, we conduct extensive experiments to evaluate \method and answer the following research questions.
\squishlist
\item \textbf{RQ1}: How does \method compare with state-of-the-art methods in the accuracy for few- and zero-shot classification on TAGs? 
\item \textbf{RQ2}: Do our augmentation techniques improve accuracy? 
\item \textbf{RQ3}: How efficient is \method in terms of training and inference?
\squishend

\begin{table*}[t]
\setlength\tabcolsep{2.pt}
\centering
\caption{Accuracy for few-shot node classification (mean±std). The best and runner-up are marked with bold and underlined, respectively. \textit{Gain} is the relative improvement of \method over the best-performing baseline.}
\begin{tabular}{c|cc|cc|cc|cc|cc}
\toprule
\multirow{2}{*}{\textbf{Method}} & \multicolumn{2}{c|}{\textbf{Cora}} & \multicolumn{2}{c|}{\textbf{Fitness}}         & \multicolumn{2}{c|}{\textbf{M.I.}}         & \multicolumn{2}{c|}{\textbf{Industrial}}   & \multicolumn{2}{c}{\textbf{Art}}          \\ \cline{2-11}
                                 & ACC                 & F1                  & ACC                  & F1                   & ACC                 & F1                  & ACC                 & F1                  & ACC                 & F1                  \\
\midrule
GCN                              & 41.15±2.41          & 34.50±2.23          & 21.64±1.34           & 12.31±1.18           & 22.54±0.82          & 16.26±0.72          & 21.08±0.45          & 15.23±0.29          & 22.47±1.78          & 15.45±1.14          \\
SAGEsup                          & 41.42±2.90          & 35.14±2.14          & 23.92±0.55           & 13.66±0.94           & 22.14±0.80          & 16.69±0.62          & 20.74±0.91          & 15.31±0.37          & 22.60±0.56          & 16.01±0.28          \\
TextGCN                          & 59.78±1.88          & 55.85±1.50          & 41.49±0.63           & 35.09±0.67           & 46.26±0.91          & 38.75±0.78          & 53.60±0.70          & 45.97±0.49          & 43.47±1.02          & 32.20±1.30          \\
\midrule
GPT-GNN                          & 76.23±1.80          & 72.23±1.17          & 48.40±0.65           & 41.86±0.89           & 67.97±2.49          & 59.89±2.51          & 62.13±0.65          & 54.47±0.67          & 65.15±1.37          & 52.79±0.83          \\
DGI                              & 78.53±1.12          & 74.58±1.24          & 47.56±0.59           & 41.98±0.77           & 68.06±0.73          & 60.64±0.61          & 52.29±0.66          & 45.26±0.51          & 65.41±0.86          & 53.57±0.75          \\
SAGEself                         & 76.32±1.25          & 73.47±1.53          & 48.90±0.80           & 41.31±0.71           & 76.70±0.48          & 70.87±0.59          & 71.87±0.61          & 65.09±0.47          & 76.13±0.94          & 65.25±0.31          \\
\midrule
GPPT                             & 75.25±1.66          & 71.16.±1.13         & 50.68±0.95           & 44.13±1.36           & 71.21±0.78          & 54.73±0.62          & 75.05±0.36          & 69.59±0.88          & 75.85±1.21          & 65.12±0.83          \\
GFP                              & 75.33±1.17          & 70.78±1.62          & 48.61±1.03           & 42.13±1.53           & 70.26±0.75          & 54.67±0.64          & 74.76±0.37          & 68.55±0.29          & 73.60±0.83          & 63.05±1.61          \\
GraphPrompt                      & 76.61±1.89          & 72.49±1.81          & 54.04±1.10           & 47.40±1.97           & 71.77±0.83          & 55.12±1.03          & 75.92±0.55          & 70.21±0.28          & 76.74±0.82          & 66.01±0.93          \\
\midrule
BERT                             & 37.86±5.31          & 32.78±5.01          & 43.26±1.25           & 34.97±1.58           & 50.14±0.68          & 42.96±1.02          & 54.00±0.20          & 47.57±0.50          & 46.39±1.05          & 37.07±0.68          \\
RoBERTa                          & 62.10±2.77          & 57.21±2.51          & 59.06±1.90           & 50.68±1.06           & 70.67±0.87          & 63.50±1.11          & 76.35±0.65          & 70.49±0.59          & 72.95±1.75          & 62.25±1.33          \\
P-Tuning v2                      & 71.00±2.03          & 66.76±1.95          & 62.12±2.92           & 52.98±2.18           & 72.08±0.51          & 65.44±0.63          & 79.65±0.38          & 74.33±0.37          & 76.86±0.59          & 66.89±1.14          \\
\midrule
G2P2                             & {\ul 80.08±1.33}    & {\ul 75.91±1.39}    & {\ul 68.24±0.53}     & {\ul 58.35±0.35}     & {\ul 82.74±1.98}    & {\ul 76.10±1.59}    & {\ul 82.40±0.90}    & {\ul 76.32±1.04}    & {\ul 81.13±1.06}    & {\ul 69.48±0.15}    \\
\midrule
\method                            & \textbf{82.66±0.77} & \textbf{79.05±1.25} & \textbf{70.79±1.09}  & \textbf{62.72±1.21}  & \textbf{87.99±0.64} & \textbf{82.61±0.81} & \textbf{85.75±0.31} & \textbf{80.45±0.25} & \textbf{85.55±0.58} & \textbf{75.59±0.16} \\
Gain                             & +3.2\%              & +4.1\%              & +3.7\%               & +7.5\%               & +6.3\%              & +8.6\%              & +4.3\%              & +5.4\%              & +5.4\%              & +8.8\%        \\     
\bottomrule
\end{tabular}
\label{tab:few-shot classification}
\end{table*}

\begin{table*}[t]
\setlength\tabcolsep{2.5pt}
\centering
\caption{Accuracy for zero-shot node classification (mean±std). The best and runner-up are marked with bold and underlined, respectively. \textit{Gain} is the relative improvement of \method over the best-performing baseline.}
\begin{tabular}{c|cc|cc|cc|cc|cc}
\toprule
\multirow{2}{*}{\textbf{Method}} & \multicolumn{2}{c|}{\textbf{Cora}} & \multicolumn{2}{c|}{\textbf{Fitness}}         & \multicolumn{2}{c|}{\textbf{M.I.}}         & \multicolumn{2}{c|}{\textbf{Industrial}}   & \multicolumn{2}{c}{\textbf{Art}}          \\ \cline{2-11}
                                 & ACC                 & F1                  & ACC                  & F1                   & ACC                 & F1                  & ACC                 & F1                  & ACC                 & F1                  \\
\midrule
BERT                             & 23.56±1.48          & 17.92±0.86          & 32.63±1.24           & 26.58±1.21           & 37.42±0.67          & 30.73±0.93          & 36.88±0.56          & 29.46±1.12          & 35.72±1.59          & 24.10±1.06          \\
BERT*                            & 23.27±1.88          & 17.27±1.92          & 34.23±1.84           & 28.20±1.73           & 50.22±0.72          & 43.34±0.78          & 55.92±2.01          & 48.46±1.27          & 55.63±1.59          & 44.12±1.02          \\
RoBERTa                          & 30.43±2.36          & 24.92±0.87          & 33.08±1.16           & 27.94±1.86           & 36.42±1.20          & 28.25±0.43          & 42.99±1.20          & 35.51±1.39          & 47.92±1.19          & 36.62±1.15          \\
RoBERTa*                         & 39.64±1.24          & 34.67±1.16          & 38.39±1.08           & 32.74±1.28           & 32.13±0.74          & 25.12±0.67          & 37.84±0.74          & 30.27±0.92          & 38.81±0.56          & 26.35±1.84          \\
G2P2                             & {\ul 64.35±2.78}    & {\ul 58.42±1.59}    & {\ul 45.99±0.69}     & {\ul 40.06±1.35}     & {\ul 74.77±1.98}    & {\ul 67.10±1.59}    & {\ul 75.66±1.42}    & {\ul 68.27±1.31}    & {\ul 75.84±1.57}    & {\ul 63.59±1.62}    \\
\midrule
Hound                            & \textbf{69.21±1.35} & \textbf{61.41±1.82} & \textbf{54.41±1.10}  & \textbf{47.45±1.63}  & \textbf{79.85±1.35} & \textbf{72.58±0.79} & \textbf{81.99±0.58} & \textbf{73.84±0.33} & \textbf{78.22±1.70} & \textbf{67.71±0.02} \\
Gain                             & +7.6\%              & +5.1\%              & +18.3\%              & +18.4\%              & +6.8\%              & +8.2\%              & +8.4\%              & +8.2\%              & +3.1\%              & +6.5\%          \\   
\bottomrule
\end{tabular}
\label{tab:zero-shot classification}
\end{table*}

\subsection{Experiment Settings}
\stitle{Datasets.} Following related researches~\cite{ni2019justifying,yan2023comprehensive}, we use the 5 datasets in Table~\ref{tab:dataset statistic} for experiments.  Cora~\cite{mccallum2000automating} is a citation network, where papers are linked by citation relations and abstract serves as the text. Art, Industrial, M.I., and Sports are derived from Amazon product categories~\cite{yan2023comprehensive}, namely, arts, crafts and sewing for Art; industrial and scientific for Industrial; musical instruments for M.I.; and sports-fitness for Fitness, respectively. For the four datasets, an edge is added to construct the graph if a user visits two products successively, and the text is the product description. The five datasets cover different scales (from thousands to millions of nodes) and number of classes (from tens to thousands).

\stitle{Baselines.} We compare \method with 13 baselines from 5 categories. 
\begin{itemize}[leftmargin=*]
    \item \textbf{End-to-end GNNs}: GCN~\cite{kipf2016semi}, SAGEsup~\cite{hamilton2017inductive}, TextGCN~\cite{yao2019graph}.  They are trained in a supervised manner for the classification tasks.
    \item \textbf{Pre-trained GNNs}: GPT-GNN~\cite{hu2020gpt}, DGI~\cite{velivckovic2018deep}, SAGEself~\cite{hamilton2017inductive}. They are first pre-trained via self-supervise learning and then fine-tuned for the classification tasks.
    \item \textbf{Graph prompt methods}: GPPT~\cite{sun2022gppt}, GFP~\cite{fang2024universal}, GraphPromt~\cite{liu2023graphprompt}. They reduce the divergence between the pre-training and downstream tasks by designing the training objectives and prompts.
    \item \textbf{Language models}: BERT~\cite{devlin2018bert}, RoBERTa~\cite{liu2019roberta}, P-Tuning v2~\cite{liu2021p}. They are first pre-trained and then fine-tuned on the text.
    \item \textbf{Co-trained model}: G2P2~\cite{wen2023augmenting}. It employs the contrastive loss to train the GNN and language model jointly such that they produce similar embeddings for node-text pairs.
\end{itemize}


Following G2P2~\cite{wen2023augmenting}, we use classification accuracy and F1 score to measure performance. We report the average value and standard deviation across 5 runs. Note that we only select language models and G2P2 as the baselines for zero-shot classification, since the other baselines require at least one labeled sample per class for either training or fine-tuning.

\stitle{Task configurations.} For few-shot classification, we use a 5-way 5-shot setup, i.e., 5 classes are taken from all classes, and then 5 nodes are sampled from these classes to construct the training set. The validation set is generated in the same way as the training set, and all remaining data is used as the test set. For zero-shot classification, we use 5-way classification, which samples classes but does not provide labeled nodes.

\begin{table*}[t]
\setlength\tabcolsep{1.5pt}
\centering
\caption{Classification accuracy for different combinations of our augmentation techniques. $\mathcal{L}_{CL}$ is the baseline that uses contrastive loss, \textit{TM} for text matching, \textit{NP} for node perturbation, and \textit{SN} for semantics negation. Best in \textit{bold}.}
\begin{tabular}{l|ccccc|ccccc}
\toprule
\multirow{2}{*}{\textbf{ }} & \multicolumn{5}{c|}{\textbf{Few-shot}} & \multicolumn{5}{c}{\textbf{Zero-shot}} 
\\ \cline{2-11}
  & \textbf{Cora}       & \textbf{Fitness}       & \textbf{M.I.}   & \textbf{Industrial}               & \textbf{Art}  & \textbf{Cora}  & \textbf{Fitness}        & \textbf{M.I.}     & \textbf{Industrial}                  & \textbf{Art}  \\
\midrule
$\mathcal{L}_{CL}$     & 80.08±1.33                           & 68.24±0.53 & 82.74±1.98 & 82.40±0.90 & 81.13±1.06                           & 64.35±2.78                           & 45.99±0.69                           & 74.77±1.98                           & 75.66±1.42                           & 75.84±1.57                           \\
$\mathcal{L}_{CL+TM}$    & 82.18±1.01                           & \textbf{70.81±1.28} & 87.91±0.59 & \textbf{85.75±0.31} & 85.37±0.60                           & 66.72±1.17                           & 49.53±3.23                           & 74.55±1.52                           & 79.79±0.59                           & 79.31±1.60                           \\
$\mathcal{L}_{CL+NP}$        &\textbf{82.66±0.77} & 70.41±3.95 & \textbf{87.99±0.64} & 85.64±0.31 & \textbf{85.55±0.58} & 68.06±1.25                           & 51.62±2.90                           & 75.43±1.08                           & 75.43±0.08                           & 78.24±1.24                           \\

$\mathcal{L}_{CL+TM+NP}$  & 82.28±0.86                           & 70.34±4.12 & 87.92±0.57 & 85.63±0.39 & 85.50±0.52                           & 66.80±1.09                           & 52.38±3.42                           & 75.83±0.86                           & 79.41±0.86                           & 78.09±1.45                           \\
\midrule
$\mathcal{L}_{CL+SN}$       & 81.18±1.49                           & 68.70±3.66 & 87.55±0.47 & 85.03±0.42 & 84.29±0.48                           & 68.09±1.38                           & 53.52±3.96                           & 78.94±1.40                           & 81.32±0.61                           & 78.37±1.69                           \\

$\mathcal{L}_{CL+TM+SN}$     & 81.51±1.31                           & 69.23±3.17 & 87.60±0.62 & 85.38±0.40 & 85.05±0.60                           & 68.60±1.22                           & 54.04±2.76                           & \textbf{79.85±1.35} & 81.85±0.55                           & 79.48±1.88                           \\
$\mathcal{L}_{CL+NP+SN}$       & 81.88±1.29                           & 69.59±3.71 & 87.88±0.46 & 85.46±0.36 & 85.17±0.73                           & \textbf{69.21±1.35} & \textbf{54.45±1.26} & 79.15±1.35                           & \textbf{81.99±0.58} & \textbf{80.22±1.70} \\
$\mathcal{L}_{CL+TM+NP+SN}$       & 81.52±1.27                           & 68.55±3.28 & 87.70±0.53 & 85.37±0.42 & 85.09±0.53                           & 68.70±1.21                           & 53.09±0.03                           & 78.88±1.28                           & 81.82±0.51                           & 80.12±1.67                          \\
\midrule
\end{tabular}
\label{tab:loss combination}
\vspace{-0.3cm}
\end{table*}

\subsection{Main Results (RQ1)}
\stitle{Few-shot node classification.} Table~\ref{tab:few-shot classification} reports the accuracy of \method and the baselines for few-shot node classification. The results show that \method consistently outperforms all baselines across the datasets, with an average improvement of 4.6\% and 6.9\% for classification accuracy and F1 score, respectively. Moreover, the improvements of \method over the baselines are over than 5\% in 6 out of the 10 cases. On Cora, the improvements of \method are smaller than the other datasets because Cora is the smallest among the datasets, and thus existing methods learn relatively well.

Regarding the baselines, end-to-end GNNs have the lowest accuracy since they are trained with only a few labeled nodes. Pre-trained GNNs outperform end-to-end GNNs because they employ self-supervised pre-training~\cite{you2020graph,zhu2021graph}, suggesting that supervision signals are important.
Graph prompt methods only utilize the graph structures and neglect the text descriptions. Conversely, language models only use the text descriptions and ignore the graph structures. G2P2~\cite{wen2023augmenting} jointly trains the GNN and language model using both the graph structures and text descriptions, and thus it achieves the best performance among the baselines. Nonetheless, \method outperforms G2P2 because it introduces more supervision signals with our augmentation tecnqiues, which help to generate more robust and informative embeddings.

\stitle{Zero-shot node classification.} Table~\ref{tab:zero-shot classification} reports the accuracy of \method and the baselines for zero-shot node classification. We only include the language models and G2P2 because the other methods require at least one labeled sample for each class. We also enhance BERT and RoBERTa as BERT* and RoBERTa* by re-tuning them on the text descriptions of the evaluated datasets (i.e., the datasets in Table~\ref{tab:dataset statistic}) to tackle domain mismatch.

The results show that \method consistently outperform all baselines by a large margin. Compared with the best-performing baseline G2P2, the average improvements of \method in classification accuracy and F1 score are  8.8\% and 9.3\%, respectively. All methods have lower accuracy for zero-shot classification than few-shot classification because zero-shot classification does not provided labeled samples, and thus the task is more challenging. However, the improvements of \method are larger for zero-shot classification because it introduces more supervision signals for learning. 

\begin{figure}[t]
    \centering
    \includegraphics[scale=0.28]{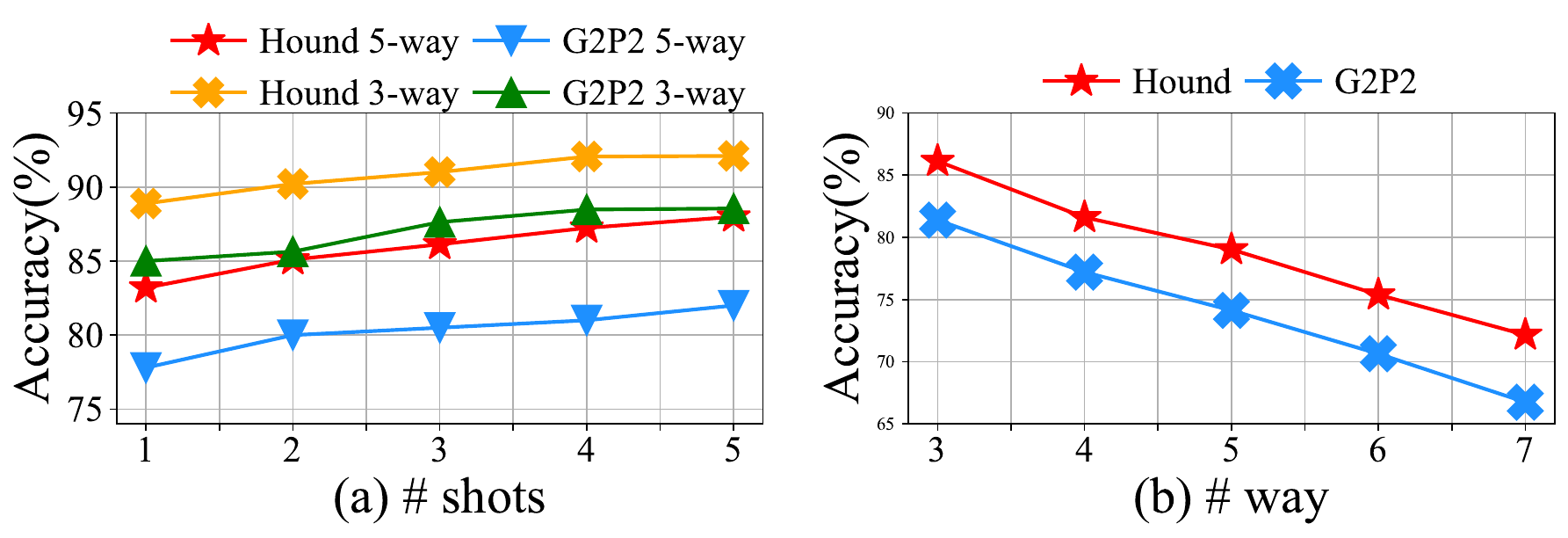}
    \caption{The accuracy comparison for our \method and G2P2 in the fewer-way and fewer-shot settings on M.I. dataset.}
    \label{fig:way-shot}
    \vspace{-0.5cm}
\end{figure}

\stitle{Robustness to task configuration.} We conduct the fewer-way and fewer-shots classification on M.I. dataset. In Figure~\ref{fig:way-shot}, we experiment with 3-way and 5-way classification and change the number of shots (i.e., number of labeled samples from each class) for few-shot classification. There are not labeled samples for zero-shot classification, thus we only change the number of ways. The results show that \method outperforms G2P2 across different configurations of ways and shots. Both \method and G2P2 perform better for 3-way classification than 5-way classification because 3-way classification is easier. Moreover, accuracy improves with the number of shots because there are more supervision signals with more labeled samples. We observe that to achieve the same accuracy, \method requires fewer labeled samples (i.e., shots) than G2P2, which helps to reduce labeling costs in practice. We also observe that \method surpasses the G2P2 across the different ways for zero-shot classification. The accuracy of both \method and G2P2 decreases with the number of way as the difficulty of classification increases.

\begin{figure}[t]
    \centering
    \includegraphics[scale=0.28]{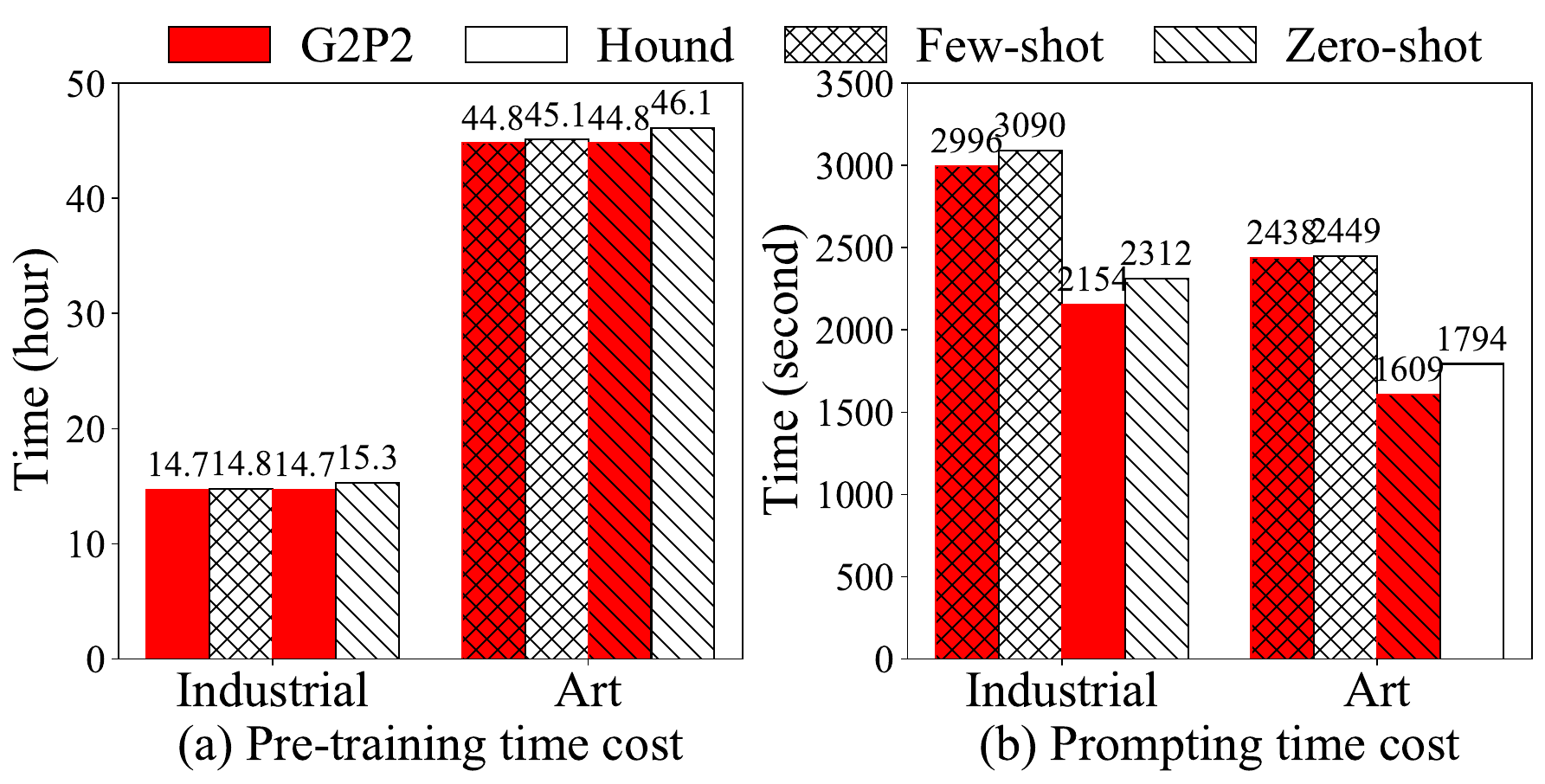}
    \caption{The time cost comparison of pre-training and prompting for G2P2 and \method.}
    \label{fig:time}
    \vspace{-0.5cm}
\end{figure}

\subsection{Micro Experiments}
\label{sec:ablation study}
\stitle{Effect of the augmentations (RQ2).} In Table~\ref{tab:loss combination}, we conduct an ablation study by trying different combinations of our three augmentations techniques. We make the following observations. 
\begin{itemize}[leftmargin=*]
	\item The augmentations are all effective in improving accuracy, as adding each of them individually outperforms the baseline $\mathcal{L}_{CL}$.
	\item The best-performing combination for few-shot classification deactivates semantic negation while zero-shot classification actives semantics negation. This is because few-shot classification uses labeled samples to learn the prompt, and the negative prompt learned by semantics negation may interfere with prompt tuning. In contrast, zero-shot classification lacks labeled data for prompt tuning, and the negative prompt helps to provide more supervision signals and improve robustness.  
	\item Text matching and node perturbation should not be utilized jointly. This may be because using both of them introduces too many node-text pairs (i.e., the perturbed embeddings of a node should be similar to multiple texts), and some of these pairs may not benefit model training. It depends on the dataset and task to decide which of them is more beneficial.        
\end{itemize}

\stitle{Efficiency (RQ3).} To examine the efficiency of \method, we compare with G2P2 for pre-training time and prompting time at inference time. We experiment on Industrial and Art, the two largest datasets, as the running time is shorter on the smaller datasets. The results in Figure~\ref{fig:time} show that \method and G2P2 have similar pre-training time and prompting time. This is because they both jointly train the GNN and language model, and computing the loss terms has a small cost compared with computing the two models. Thus, even though \method uses more loss terms, the additional overheads is negligible. Few-shot classification has longer prompting time than zero-shot classification because it needs to tune the prompt using the labeled samples.

\begin{figure}[t]
    \centering
    \includegraphics[scale=0.28]{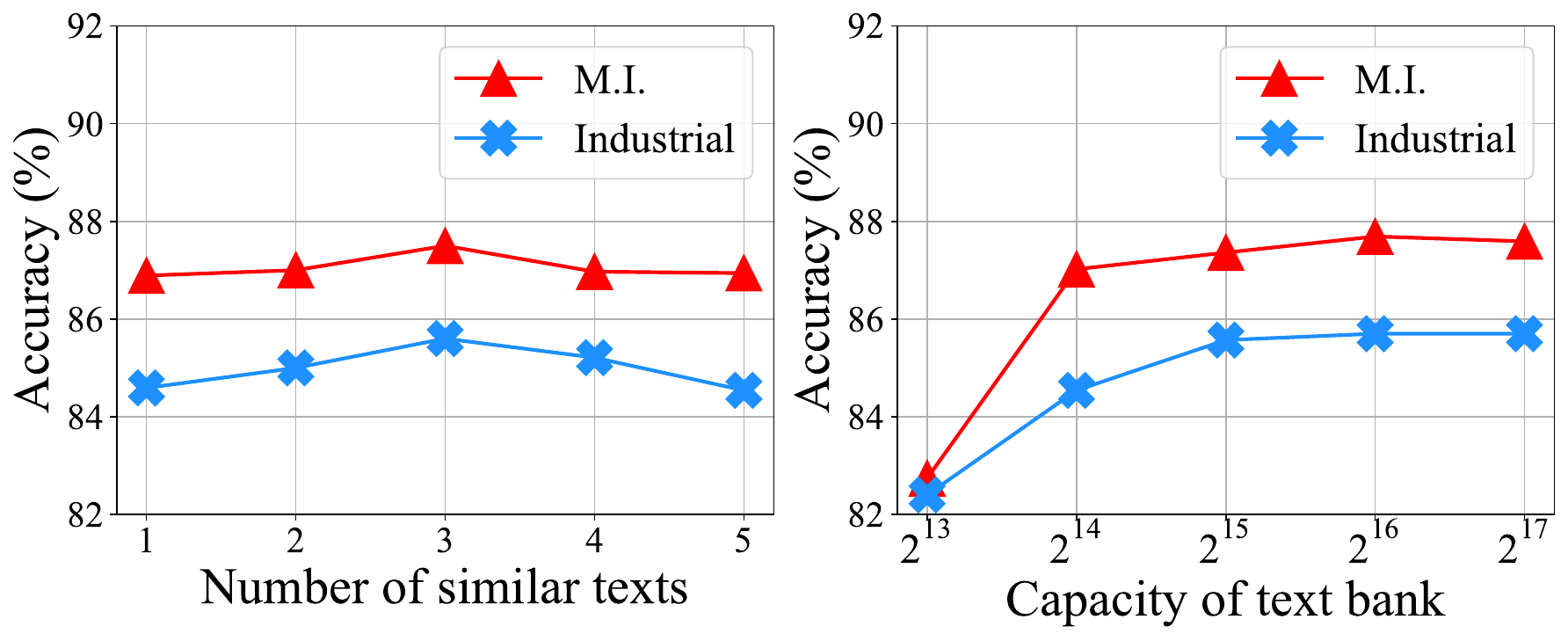}
    \caption{The comparison of the number of similar texts and the capacity of text bank for \method on M.I. anf Industrial.}
    \label{fig:text bank}
    
\end{figure}

\stitle{Parameters.} Recall the text matching has two parameters, i.e., the number of similar texts for each node and the capacity of text bank. Figure~\ref{fig:text bank} examines the effect of the two parameters on the Industrial and M.I. datasets. Note that text matching is disabled when the capacity of text bank is zero. We observe that accuracy first increases but then decreases with the number of similar texts. This is because while more similar texts can provide more supervision signals, an excessive number of these signals may introduce noise by including texts that are not truly similar to the target node. Hence, the optimal accuracy are obtained at an intermediate value to balance between supervision signals and noises. When increasing the capacity of the text bank, accuracy first increases but then stabilizes. This is because using a larger text bank allows a node to identify texts that are more similar but the similarity will become sufficiently highly when the bank is large enough.

\section{Related Work}
\stitle{Graph Pre-training and Prompting.}  GNNs~\cite{kipf2016semi,velivckovic2017graph,xu2018powerful,wu2019simplifying,kipf2016variational} use message passing to aggregate features from neighboring nodes to compute graph node embedding. However, early GNN models, such as  GCN~\cite{kipf2016semi}, GIN~\cite{xu2018powerful}, and GAT~\cite{velivckovic2017graph}, are supervised and require many labeled nodes for training. To mine supervision signals from unlabeled data, graph self-supervised learning is proposed to train using well-designed pretext tasks~\cite{zhu2021graph,you2020graph,hamilton2017inductive,velivckovic2018deep,hu2020gpt}. For instance, DGI~\cite{velivckovic2018deep} learns node embeddings by maximizing mutual information between the global and local node embeddings.  GPT-GNN~\cite{hu2020gpt} utilizes a self-supervised graph generation task to combine the graph structural and semantic information. GraphMAE~\cite{hou2022graphmae} learns robust graph node embeddings by masking graph nodes or edges and then reconstructing them.

Graph self-supervised learning methods still require many labeled instances to fine-tune specific tasks (e.g., node classification). To further reduce the reliance on labeled instances, graph prompt learning~\cite{sun2022gppt,fang2024universal,liu2023graphprompt,sun2023all} is proposed for few-shot node classification. For example, GPPT predicts the node label by deciding whether an edge exists between the target node and candidate labels. GFP~\cite{fang2024universal} learns a parameterized feature as a prompt to be added to the original node features. GraphPrompt~\cite{liu2023graphprompt} learns embeddings for subgraphs rather than nodes to unify graph-level and node-level tasks. These approaches consider only the graph and thus have limited accuracy for TAGs with text descriptions. To account for the text, TextGCN~\cite{yao2019graph} generates text embeddings using pre-trained language models and adds these embeddings as node features for GNN training. G2P2~\cite{wen2023augmenting} jointly trains the language model and GNN with the contrastive strategy and uses prompting for few-shot and zero-shot node classification. 

Like graph pre-training methods, \method designs pre-training tasks to learn from unlabeled data. However,  \method targets TAGs and considers the graph and text modalities jointly by mining more node-text pairs for training while graph pre-training methods consider only the graph (e.g., by using node pairs or subgraphs).

\stitle{Pre-trained Language Models (PLMs).} PLMs~\cite{devlin2018bert,liu2019roberta,liu2021p,raffel2020exploring,brown2020language,yang2019xlnet} enhance the ability to understand and generate natural language by pre-training on large-scale text corpus. The well-known BERT~\cite{devlin2018bert}, for instance, is pre-trained with two tasks, i.e., masked token reconstruction and next token prediction, to capture contextual information. RoBERTa~\cite{liu2019roberta} improves BERT by eliminating the next token prediction task, increasing the batch size and data volume during pre-training, and using a dynamic masking strategy. P-Tuning v2~\cite{liu2021p} introduces learnable prompts to a pre-trained model's inputs to guide the model to focus on task relevant information. While PLMs achieve great success for text oriented tasks, they cannot capture the topology information for TAGs.

\section{Conclusion}
In this paper, we study few-shot and zero-shot node classification on text-attributed graphs. We observe that the accuracy of existing methods is unsatisfactory due to the lack of supervision signals, and propose \method as a novel pre-training and prompting framework to enhance supervision. \method  incorporates three key augmentation techniques, i.e., node perturbation, text matching, and semantics negation, to mine supervision signals from both the graph and text modalities. Extensive experiments show that \method outperforms existing methods by a large margin. We think \method's methodology, i.e., generating node-text pairs that should have similar/dissimilar embeddings to enforce priors, is general and can be extended beyond our augmentation techniques.


\clearpage
\bibliographystyle{ACM-Reference-Format}
\balance
\bibliography{ref}










\end{document}